\documentclass[runningheads]{llncs}

\usepackage[utf8]{inputenc}
\usepackage[T1]{fontenc}
\usepackage{lmodern}
\usepackage{graphicx}
\usepackage{amsfonts,amssymb,here,dsfont,hyperref}
\usepackage{amsmath}               
\newtheorem{assumption}{Assumption}
\usepackage{color}
\usepackage{version}
\usepackage{todonotes}
\usepackage[lofdepth,lotdepth]{subfig}

\usepackage{wrapfig}
\usepackage{xcolor}
\usepackage{multirow}
\usepackage{nicefrac}       
\usepackage{microtype}      
\usepackage{algorithm, algorithmic}
\newsubfloat{table}
\usepackage{multirow} 
\usepackage{diagbox}
\usepackage{booktabs}       
\usepackage{colortbl}
\usepackage{caption}

\usepackage{diagbox}
\usepackage{cite}
\usepackage[makeroom]{cancel}
\usepackage[normalem]{ulem}

\newcommand{\qedsymbol}{$\blacksquare$}
\newcommand{\indep}{\perp \!\!\! \perp}
\newcommand{\one}{\mathds{1}}

\DeclareMathOperator*{\argmin}{argmin}

\begin{document}

\toctitle{Fairness in Multi-Task Learning via Wasserstein Barycenters}
\tocauthor{Fran\c{c}ois~Hu, Philipp~Ratz and Arthur~Charpentier}

\title{Fairness in Multi-Task Learning\\via Wasserstein Barycenters}

\author{Fran\c{c}ois Hu\inst{1}\orcidID{0009-0000-6093-6175} \and
Philipp Ratz\inst{2}\orcidID{0000-0002-0966-5493} \and
Arthur Charpentier\inst{2}\orcidID{0000-0003-3654-6286}}

\authorrunning{F. Hu, P. Ratz and A. Charpentier}

\institute{Université de Montréal, Montréal, Québec, Canada 
\email{francois.hu@umontreal.ca}\and
Université du Québec à Montréal, Montréal, Québec, Canada
\email{ratz.philipp@courrier.uqam.ca}, \email{charpentier.arthur@uqam.ca}}

\maketitle

\begin{abstract}
Algorithmic Fairness is an established field in machine learning that aims to reduce biases in data. Recent advances have proposed various methods to ensure fairness in a univariate environment, where the goal is to de-bias a single task. However, extending fairness to a multi-task setting, where more than one objective is optimised using a shared representation, remains underexplored. To bridge this gap, we develop a method that extends the definition of \textit{Strong Demographic Parity} to multi-task learning using multi-marginal Wasserstein barycenters. Our approach provides a closed form solution for the optimal fair multi-task predictor including both regression and binary classification tasks. We develop a data-driven estimation procedure for the solution and run numerical experiments on both synthetic and real datasets. The empirical results highlight the practical value of our post-processing methodology in promoting fair decision-making.


\end{abstract}
%
%
%
    

\section{Introduction}


Multi-task learning (MTL) is a loosely defined field that aims to improve model performance by taking advantage of similarities between related estimation problems through a common representation \cite{ruder2017overview, zhang2021survey}. MTL has gained traction in recent years, as it can avoid over-fitting and improve generalisation for task-specific models, while at the same time being computationally more efficient than training separate models\cite{baxter2000model}. For these reasons, the usage of MTL is likely to grow and spread to more disciplines, thus ensuring fairness in this setting becomes essential to overcome historical bias and prevent unwanted discrimination. Indeed, in many industries, discriminating on a series of sensitive features is even prohibited by law \cite{ecj_case}. Despite the apparent importance of fairness, it remains challenging to incorporate fairness constraints into MTL due to its multivariate nature.

Algorithmic fairness refers to the challenge of reducing the influence of a sensitive attribute on a set of predictions. With increased model complexity, simply excluding the sensitive features in the model is not sufficient, as complex models can simply proxy for omitted variables. Several notions of fairness have been considered~\cite{barocas-hardt-narayanan,zafar2019fairness} in the literature. In this paper, we focus on the \emph{Demographic Parity} (DP)~\cite{calders2009building} that requires the independence between the sensitive feature and the predictions, while not relying on labels (for other notions of fairness, see \emph{Equality of odds} or \emph{Equal opportunity}~\cite{hardt2016equality}). This choice is quite restrictive in the applications, but provides a first stepping stone to extend our findings to other definitions. In single-task learning problems, the fairness constraint (such as DP) has been widely studied for classification or regression \cite{calders2009building, zemel2013learning, zafar2017fairness, Chzhen_Denis_Hebiri_Oneto_Pontil19, agarwal2019fair, denis2021fairness}, but to extend fairness to multiple tasks, we first need to study the effects of learning tasks jointly on the potential outcomes. In line with a core advantage of MTL, the approach we propose is based on post-processing which results in faster computations than other approaches discussed below. The contributions of the present article can hence be summarised as follows:

\subsubsection{Contributions} We consider multi-task problems that combine regression and binary classification, with the goal of producing a fair shared representation under the DP fairness constraint. More specifically:
\begin{itemize}
    \item We transform the multi-task problem under Demographic Parity fairness to the construction of multi-marginal Wasserstein-2 barycenters. Notably, we propose a closed form solution for the optimal fair multi-task predictor.
    \item Based on this optimal solution, we build a standard data-driven approach that mimics the performance of the optimal predictor both in terms of risk and fairness. In particular, our method is post-processing and can be applied to any off-the-shelf estimators. 
    \item Our approach is numerically illustrated on several real data sets and proves to be very efficient in reducing unfairness while maintaining the advantages of multi-task learning.
\end{itemize}

\subsubsection{Related work}

Algorithmic fairness can be categorised into: 1) \textit{pre-processing} methods which enforce fairness in the data before applying machine learning models \cite{calmon2017optimized, adebayo2016iterative, plevcko2020fair}; 2) \textit{in-processing} methods, who achieve fairness in the training step of the learning model \cite{Agarwal_Beygelzimer_Dubik_Langford_Wallach18,Donini_Oneto_Ben-David_Taylor_Pontil18,agarwal2019fair}; 3) \textit{post-processing} which reduces unfairness in the model inferences following the learning procedure\cite{chiappa2020general,Chzhen_Denis_Hebiri_Oneto_Pontil20Wasser, Chzhen_Denis_Hebiri_Oneto_Pontil20Recali}. Our work falls into the latter. This comes with several computational advantages, not least the fact that even partially pre-trained models can be made fair, which extends our findings to multi-task transfer learning.

Within standard, single-task classification or regression problems, the DP constraint has been extensively studied before. In particular, the problem of integrating algorithmic fairness with the Wasserstein distance based barycenter has been an active area of research \cite{chiappa2020general, gordaliza2019obtaining, pmlr-v115-jiang20a, Chzhen_Denis_Hebiri_Oneto_Pontil20Wasser} but most studies focus on learning univariate fair functions. Our work differs from the aforementioned work by enforcing the DP-fairness in multi-task learning, involving learning a fair vector-valued function based on a shared representation function. To the best of our knowledge, there is only a limited body of research concerning fairness in MTL settings. For instance, Zhao et al. \cite{zhao2019rank} introduced a method for fair multi-task regression problems using rank-based loss functions to ensure DP-fairness, while \cite{roy2023learning} and \cite{wang2021understanding} independently achieve fairness for multi-task classification problems in the Equal Opportunity or Equalised Odds sense. However, our approach offers a flexible framework for achieving fairness by simultaneously training fair predictors including binary classification and regression. Oneto et al. \cite{oneto2020learning, oneto2020exploiting} suggested a DP-fair multi-task learning approach that learns predictors using information from different groups. They proposed this for linear~\cite{oneto2020learning} and 1-hidden layer networks~\cite{oneto2020exploiting} predictors. Our work extends this approach to arbitrary multivariate distributions and proposes a post-processing method that keeps additional computations to a minimum.

\subsubsection{Outline of the paper} The remainder of this article is structured as follows: Section \ref{sec:PbStatement} introduces MTL, DP-fairness and the objective in rendering multi-task problems fair. Section \ref{sec:wasserstein} introduces our fair multi-task predictor which is then translated to an empirical plug-in estimator in Section \ref{sec:plugin}. Section \ref{sec:application} evaluates the estimator on synthetic and real data and we conclude in Section \ref{sec:conclusion}.

\section{Problem Statement}\label{sec:PbStatement}

In machine learning, one often encounters two types of prediction tasks: regression and binary classification. In regression, the goal is to predict a real-valued output in $\mathbb{R}$ while in binary classification, the goal is to predict one of two classes $\{0, 1\}$. Although the definitions and our approach can be applied to any number of finite tasks, for ease of presentation we focus this section on these two sub-cases. 

\subsection{Multi-Task Learning}\label{sec:mtr}

There are several definitions and goals that can be achieved through MTL. As our applications are centered on similar tasks, we focus on one aspect referred to as \emph{parameter sharing} between the tasks (for a more comprehensive survey, we recommend Zhang and Yang's survey \cite{zhang2021survey}). Parameter sharing is especially useful in the case where there are missing labels in one of the tasks, as MTL can exploit similarities among the tasks to improve the predictive performance. Formally, we let $(\boldsymbol{X},S,\boldsymbol{Y})$ be a random tuple with distribution $\mathbb{P}$. Here, $\boldsymbol{X}$ represents the non-sensitive features, $S$ a sensitive feature, considered discrete, across which we would like to impose fairness and $\boldsymbol{Y}$ represents the tasks to be estimated. In theory, there are no restrictions on the space of $\boldsymbol{X}$, $\boldsymbol{Y}$, or $S$. Throughout the article, to ease the notational load, we assume that $\boldsymbol{X} \in \cal{X}\subset$ $\mathbb{R}^{d}$, $\mathcal{S} = \{-1, 1\}$ where $-1$ represents the minority group and $1$ the majority group and $\boldsymbol{Y} = (Y_1, Y_2)  \in \mathcal{Y}_1\times \mathcal{Y}_2$ where  $\mathcal{Y}_1\subset \mathbb{R}$ and $\mathcal{Y}_2 =\lbrace 0, 1 \rbrace$ (or $[0, 1]$ if we consider \textit{score} function). That is, we consider problems where the columns of $\boldsymbol{Y}$ represent regression-binary classification problems. More specifically, we consider for $g_1^{*} : \mathcal{X}\times\mathcal{S} \to \mathbb{R}$ the general regression problem 
\begin{equation}\label{eq:RegGeneral}
   Y_1 = g_1^{*}(\boldsymbol{X}, S) + \zeta 
\end{equation}
with $\zeta\in\mathbb{R}$ a zero mean noise. $g_1^{*}$ is the regression function that minimises the squared risk $\mathcal{R}_{L_2}(g) := \mathbb{E}\left(Y_1 - g(\boldsymbol{X}, S) \right)^2$. For the second task, recall that a classification rule $c_2:\mathcal{X}\times\mathcal{S}\to\{0, 1\}$ is a function evaluated through the misclassification risk $\mathcal{R}_{0-1}(c) := \mathbb{P}\left(c(\boldsymbol{X},S) \neq Y_2\right)$. We denote $g_2^*(\boldsymbol{X},S) := \mathbb{P}(Y_2 = 1 | \boldsymbol{X}, S)$ the conditional probability (or score) of belonging to class $1$. Recall that the minimisation of the risk $\mathcal{R}_{0-1}(\cdot)$ over the set of all classifiers is given by the Bayes classifier 
\begin{equation}\label{eq:ClassGeneral}
c_2^{*}(\boldsymbol{X},S) = \one\left\{g_2^*(\boldsymbol{X},S) \geq 1/2\right\}\enspace.
\end{equation}
The modelling of the two columns of $\boldsymbol{Y}$ is then referred to as the \emph{tasks}, denoted $\mathcal{T} = \{ 1, 2 \}$. Here we adopt the general notation the two tasks $Y_1$ and $Y_2$ are modelled on the same input space $\mathcal{X}\times \mathcal{S}$ such that they are independent of each other conditionally on $(\boldsymbol{X}, S)$. In line with the notion of related tasks, we suppose that the tasks share a common representation of the features $h_{\theta}:\mathcal{X}\times\mathcal{S}\to \mathcal{Z}$ where $\mathcal{Z}\subset \mathbb{R}^r$ and the marginal task models can be represented by $g_t(\cdot) = f_t \circ h_\theta (\cdot)$ for a given task-related function $f_t:\mathcal{Z}\to \mathcal{Y}_t$. The representation can then be approximated via a neural network. We denote $\mathcal{H}$ the set of all representation functions. To appropriately weigh each of the tasks in the estimation problem, we use trade-off weights $\boldsymbol{\lambda} = (\lambda_1, \lambda_2)$ where we assume $\lambda_t > 0$ for all $t$. 
This yields the simple multi-task estimator defined as:
 \begin{equation}\label{eq:multitask}
     \boldsymbol{\theta}_{\boldsymbol{\lambda}}^* = \argmin_{\theta} \mathbb{E} \left[\sum_{t=1}^2 \lambda_t \mathcal{R}_t\big(Y_t, f_t \circ h_{\theta}(\boldsymbol{X}, S)\big) \right]
 \end{equation}
with $\mathcal{R}_t$ the risk associated to task $t$. Restricting each task to use the same representation $h_{\theta}$ might seem overly simplistic, but given that under mild conditions the universal approximation theorem \cite{hornik1989multilayer} is applicable, a large variety of problems can still be modelled. A thorough discussion of the advantages of multi-task learning would go beyond the scope of this article and we refer the interested reader instead to \cite{zhang2021survey, ruder2017overview} for a comprehensive survey. The empirical estimation of Eq.\eqref{eq:multitask} will be further discussed in Section \ref{sec:empmultitask}. 

\subsubsection{Notations} Assuming that the following density exists, for each $s\in\mathcal{S}$ and for any task predictor $g$, we denote $\nu_g$ the probability measure of $g(\boldsymbol{X}, S)$ and $\nu_{g|s}$ the probability measure of $g(\boldsymbol{X}, S)|S=s$. $F_{g|s}:\mathbb{R}\to [0, 1]$ and $Q_{g|s}:[0, 1] \to \mathbb{R}$ are, respectively, its CDF function defined as $F_{g|s}(u) := \mathbb{P}\left( g(\boldsymbol{X}, S) \leq u|S=s \right)$ and its corresponding quantile function defined as $Q_{g|s}(v) := \inf\{u\in\mathbb{R}:F_{g|s}(u)\geq v \}$.
\begin{figure}[ht]
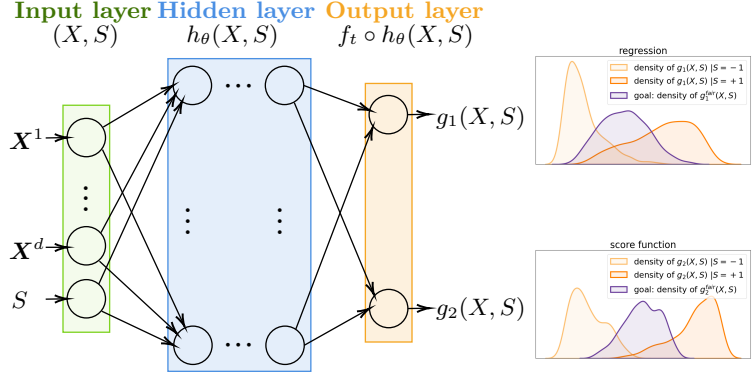

	\vspace{.1in}
	\centering
    \include{tikz/big_intro}
    \caption{Representation function sharing in a neural network for multi-task learning. The goal in DP-fairness is to construct a set of predictors $\{g^{\text{fair}}_t(\boldsymbol{X}, S)\}_t$ independent from the sensitive feature $S$. $\boldsymbol{X}^i$ refers to the $i$-th feature of $\boldsymbol{X}$.}
    \label{fig:intro}
\end{figure}
\subsection{Demographic Parity}\label{sec:fairness}




We introduce in this section the fairness under \textit{Demographic Parity} (DP) constraint in both single-task and multi-task problems.

\subsubsection{Fairness in single-task problems} 
For a given task $t\in\mathcal{T}=\{1, 2\}$, we denote by $\mathcal{G}_t$ the set of all predictors $g_t:\boldsymbol{X}\times \mathcal{S}\to \mathcal{Y}_t$  of the form $g_t(\cdot) = f_t \circ h_\theta (\cdot)$. In particular for the binary classification, $\mathcal{G}_2$ represents the set of all score functions in $\mathcal{Y}_2 = [0, 1]$ and additionally we denote $\mathcal{G}^{\text{class}}_2$ the set of all classifiers in $\{0, 1\}$. 
With a provided score function $g_2\in\mathcal{G}_2$, a class prediction $c_2\in\mathcal{G}^{\text{class}}_2$ is generated using a threshold $\tau\in [0, 1]$, expressed as $c_2(\cdot) = \one\{g_2(\cdot) \geq \tau\}$. Most work aims to ensure that sensitive information $S$ (such as \textit{race}) does not influence the decisions $c_2$, i.e. $c_2(\boldsymbol{X}, S) \indep S$. This fairness criterion is called \textit{weak} Demographic Parity \cite{hardt2016equality,lipton2018does} and verifies
$$
\left|\ \mathbb{P}(c_2(\boldsymbol{X}, S) = 1 \ | \ S = -1) - \mathbb{P}(c_2(\boldsymbol{X}, S) = 1 \ | \ S = 1)\ \right| = 0\enspace.
$$
However, enforcing DP fairness for a given threshold does not imply enforcing DP fairness for other thresholds. Therefore we need to enforce the score function $g_2$ instead, i.e. $g_2(\boldsymbol{X}, S) \indep S$. This definition, called \textit{strong} Demographic Parity~\cite{pmlr-v115-jiang20a,agarwal2019fair}, will be formally defined below in Definition~\ref{def:SDP}.



\begin{remark}[Misclassification risk and squared risk]\label{rem:BiClRisk}
    In binary task $\{0, 1\}$, given $\tau = 1/2$ the misclassification risk can be rewritten as
    $$
    \mathbb{P}\left( Y_2 \neq c_2^*(\boldsymbol{X}, S) \right) = \mathbb{E}\left[\left( Y_2 - c_2^*(\boldsymbol{X}, S) \right)^2\right]
    $$
    with $g_2^*(\boldsymbol{X}, S) = \mathbb{P}\left(Y_2=1| \boldsymbol{X}, S\right) = \mathbb{E}\left[Y_2 | \boldsymbol{X}, S\right]$.
    Since our goal is to enforce fairness w.r.t. the sensitive feature $S$ in a score function $g_2\in\mathcal{G}_2$, we are interested in minimising the risk 
    $\mathbb{E}\left( Y_2 - g_2(\boldsymbol{X}, S) \right)^2$ instead.
    Notably, for any given task $t\in\{1, 2\}$, the (unconstrained) single-task objective becomes:
    $$g_t^* \in \argmin_{g_t\in\mathcal{G}_t}\mathbb{E}\left[\left( Y_t - g_t(\boldsymbol{X}, S) \right)^2\right]
    .$$
    
\end{remark}

We now formally define the (strong) Demographic Parity notion of fairness and the associated unfairness measure.

\begin{definition}[Strong Demographic Parity]\label{def:SDP}
Given a task $t\in \mathcal{T}$ (regression or score function), a predictor $g_t:\boldsymbol{X}\times \mathcal{S} \to \mathcal{Y}_t\subset\mathbb{R}$
is called fair under Demographic Parity (or \textit{DP-fair}) if for all $s, s'\in\mathcal{S}$
\begin{equation*}
    \label{eq:DPFair}
\sup\limits_{u\in\mathcal{Y}_t} \left|\ \mathbb{P}(g_t(\boldsymbol{X}, S) \leq u \ | \ S = s) - \mathbb{P}(g_t(\boldsymbol{X}, S) \leq u \ | \ S = s')\ \right| = 0 \enspace.
\end{equation*}
\end{definition}

\begin{definition}[Unfairness]\label{def:Unfairness}
The unfairness of $g_t\in\mathcal{G}_t$ is quantified by
\begin{equation}
    \label{eq:Unfairness}
\mathcal{U}(g_t) := \max_{s, s'\in\mathcal{S}}\sup\limits_{u\in\mathcal{Y}_t} \left|\ F_{g_t|s}(u) - F_{g_t|s'}(u)\ \right| \enspace.
\end{equation}
Hence, by the above definition, a predictor $g_t$ is fair if and only if $\mathcal{U}(g_t) = 0$.
\end{definition}
We use $\mathcal{G}^{\text{fair}}_{t} := \left\{ g\in \mathcal{G}_t : g \text{ is DP-fair}\right\}$ to denote the set of DP-fair predictors in $\mathcal{Y}_t$ for a given task $t\in\mathcal{T}$. In single-task learning for regression and binary classification, the aim in DP fairness is to minimise the squared risk over $\mathcal{G}^{\text{fair}}_{t}$ to find a fair predictor
\begin{equation}\label{eq:MinDPScoreFair2}
    g_t^{*(\text{fair})} \in \argmin_{g_t \in \mathcal{G}^{\text{fair}}_{t} } \mathbb{E}\left[\left(Y_t - g_t(\boldsymbol{X}, S) \right)^2 \right]\enspace.
\end{equation}
Note that the estimator of the optimal regression for this optimisation problem~\eqref{eq:MinDPScoreFair2} can be 
identified as the solution of the Wasserstein barycenter problem~\cite{Chzhen_Denis_Hebiri_Oneto_Pontil20Wasser,gouic2020projection,pmlr-v115-jiang20a}. In binary classification, \cite{gaucher2022fair} show that maximising accuracy under DP fairness constraint is the same as solving a corresponding score function with the threshold at level $\tau = 1/2$. Here, we extend this notation as suggested in Remark \ref{rem:BiClRisk}.

\subsubsection{Fairness in multi-task problems}
Given trade-off weight $\boldsymbol{\lambda} = (\lambda_t)_{t\in\mathcal{T}}$ and multi-task problem $\boldsymbol{Y} = \left(Y_t\right)_{t\in\mathcal{T}}$, an optimal multi-task predictor takes a feature set $(\boldsymbol{X}, S)$ as input and outputs a set of predictions denoted $(g_{t, \boldsymbol{\lambda}}^*)_{t\in\mathcal{T}}$. The $t$-th marginal prediction is given by $g_{t,\boldsymbol{\lambda}}^*(\cdot) = f_t \circ h_{\theta_{\boldsymbol{\lambda}}^*}(\cdot)$. Alternatively, through a slight abuse of notation, we can express it as $g_{t,\boldsymbol{\lambda}}^*(\cdot) = f_t \circ \theta_{\boldsymbol{\lambda}}^*(\cdot)$, where the representation function yields
\begin{equation*}
\theta_{\boldsymbol{\lambda}}^* \in \argmin_{\theta \in \mathcal{H}} \ \mathbb{E}\left[\sum_{t\in\mathcal{T}}\lambda_t\left(Y_t - f_t\circ \theta(\boldsymbol{X}, S) \right)^2
\right]\enspace.
\end{equation*}
For the sake of simplicity in presentation, we will represent the function $h_\theta$ as $\theta$ from this point forward.
A multi-task predictor is DP-fair if its associated marginal predictor satisfies DP fairness in Definition~\ref{def:SDP} for every task $t\in\mathcal{T}$. We use $\mathcal{H}^{\text{fair}} := \{\theta \in\mathcal{H} : f_t \circ \theta \text{ is DP-fair for each task } t\in\mathcal{T}\}$ to denote the subset of all representations where each task is DP-constrained. The constrained multi-objective optimisation of $\boldsymbol{Y} = \left(Y_t\right)_{t\in\mathcal{T}}$ is given by the fair optimal representation function
\begin{equation}\label{eq:MainFairOptimalEq}
{\theta^{*(\text{fair})}_{\boldsymbol{\lambda}}} \in \argmin_{\substack{\theta \in \mathcal{H}^{\text{fair}}}} \ \mathbb{E}\left[\sum_{t\in\mathcal{T}}\lambda_t\left(Y_t - f_t\circ \theta(\boldsymbol{X}, S) \right)^2
\right]\enspace.
\end{equation}
Notably, for each task $t\in\mathcal{T}$, the associated marginal fair optimal predictor is naturally denoted $g_{t, \boldsymbol{\lambda}}^{*(\text{fair})}(\boldsymbol{X}, S) = f_t \circ {\theta^{*(\text{fair})}_{\boldsymbol{\lambda}}}(\boldsymbol{X}, S)$. 
$(f_1, \dots, f_{|\mathcal{T}|})$ is predetermined to match the output type of each task in $(Y_1, \dots, Y_{|\mathcal{T}|})$. For instance, one can use linear activation functions for regression problems, and sigmoid functions for binary classification.

\section{Wasserstein fair multi-task predictor}\label{sec:wasserstein}

We describe in this section our proposed post-processing approach for constructing a fair multi-task learning. To derive a characterisation of the optimal fair predictor, we work under the following assumption.

\begin{assumption}[\textbf{Continuity assumption}]\label{assu:continuity}
    For any $(s, t, \boldsymbol{\lambda})\in \mathcal{S}\times \mathcal{T}\times \Lambda$, we assume that the measure $\nu_{g^*_{t, \boldsymbol{\lambda}}|s}$ has a density function. This is equivalent to assuming that the mapping $u \mapsto F_{g^*_{t, \boldsymbol{\lambda}}|s}(u)$ is continuous.
\end{assumption}

Driven by our goal to minimise the squared risk defined in Eq.\eqref{eq:MainFairOptimalEq} and building upon previous research in the univariate case \cite{Chzhen_Denis_Hebiri_Oneto_Pontil20Wasser, gouic2020projection}, we introduce the Wasserstein-2 distance. We then demonstrate that fairness in the multi-task problem can be framed as the optimal transport problem involving the Wasserstein-2 distance. The relationship between these concepts is established in Thm.~\ref{prop:MainFairOptimal}.

\begin{definition}[Wasserstein-2 distance]\label{def:W2}
    Let $\nu$ and $\nu'$ be two univariate probability measures. The Wasserstein-2 distance between $\nu$ and $\nu'$ is defined as
    \begin{equation*}
        \mathcal{W}_2^2(\nu, \nu') = \inf_{\gamma\in\Gamma_{\nu, \nu'}} \left\{ \int_{\mathbb{R}\times\mathbb{R}} |y-y'|^2d\gamma(y, y') \right\}
    \end{equation*}
    where $\Gamma_{\nu, \nu'}$ is the set of distributions on $\mathbb{R}\times\mathbb{R}$ having $\nu$ and $\nu'$ as marginals.
\end{definition}

The proof of the following theorem is based on results from~\cite{Chzhen_Denis_Hebiri_Oneto_Pontil20Wasser} or~\cite{gouic2020projection}. Although their work is not immediately applicable to our case due to the dependence of the tasks, they provide valuable insights on the use of optimal transport theory in the context of Demographic Parity. We provide a sketch of a proof but relegate the rigorous version to the Appendix.

\begin{theorem}[Optimal fair predictions]\label{prop:MainFairOptimal} Let Assumption~\ref{assu:continuity} be satisfied. Recall that $\pi_s = \mathbb{P}(S=s)$. 

\begin{enumerate}
    \item A representation function $\theta^{*(\text{fair})}_{\boldsymbol{\lambda}}$ satisfies Eq.\eqref{eq:MainFairOptimalEq}, \textit{i.e.}, 
\begin{equation*}
        \theta^{*(\text{fair})}_{\boldsymbol{\lambda}} \in \argmin_{\substack{\theta \in \mathcal{H}^{\text{fair}}}} \mathbb{E}\left[\sum_{t\in\mathcal{T}}\lambda_t\left(Y_t - f_t\circ \theta(\boldsymbol{X}, S) \right)^2
\right]\enspace.
\end{equation*}
if and only if, for each $t\in\mathcal{T}$ this same representation function satisfies 
$$\nu_{f_t\circ \theta^{*(\text{fair})}_{\boldsymbol{\lambda}}} \in \argmin_{\nu}\sum_{s\in\mathcal{S}}\pi_s \mathcal{W}_2^2(\nu_{g_{t, \boldsymbol{\lambda}}^*|s}, \nu )\enspace.$$ 
\item Additionally, the optimal fair predictor $g^{*(\text{fair})}_{t, \boldsymbol{\lambda}} (\cdot)= f_t\circ \theta^{*(\text{fair})}_{\boldsymbol{\lambda}}(\cdot)$ can be rewritten as 
\begin{equation}\label{eq:FairPredTh}
        g^{*(\text{fair})}_{t, \boldsymbol{\lambda}}(\boldsymbol{x}, s) = \sum_{\substack{s'\in\mathcal{S} }}  \pi_{s'}Q_{g^*_{t,\boldsymbol{\lambda}}|s'}\circ F_{g^*_{t,\boldsymbol{\lambda}}|s}\left(g^*_{t,\boldsymbol{\lambda}}(\boldsymbol{x}, s)\right) ,\ \ (\boldsymbol{x}, s)\in\mathcal{X}\times\mathcal{S} \enspace.
    \end{equation}
\end{enumerate}
\end{theorem}

\begin{proof}[sketch]
    Recall Eq.\eqref{eq:RegGeneral} and $g_2^*(\boldsymbol{X}, S) = \mathbb{E}\left( Y_2 |\boldsymbol{X}, S \right)$, the multi-objective described in Eq.\eqref{eq:MainFairOptimalEq} can be easily rewritten
\begin{equation*}
\min_{\substack{\theta \in \mathcal{H}^{\text{fair}}}} \ \mathbb{E}\left[\sum_{t\in\mathcal{T}}\lambda_t\left(g_t^*(\boldsymbol{X}, S) - f_t\circ \theta(\boldsymbol{X}, S) \right)^2
\right]\enspace.
\end{equation*}
Using Prop.1 in~\cite{dosovitskiy2020you} together with A.\ref{assu:continuity}, there exists a function $V_t:\mathcal{X}\times \mathcal{S}\times \Lambda\to \mathcal{Y}_t$ (or $g_{t, \boldsymbol{\lambda}}^{*}(\boldsymbol{x}, s)$ by abuse of notation) such that the optimisation is equivalent to
\begin{equation*}
\min_{\substack{\theta \in \mathcal{H}^{\text{fair}}}} \ \mathbb{E}_{\boldsymbol{\lambda}\sim \mathbb{P}_{\boldsymbol{\lambda}}}\mathbb{E}\left[\sum_{t\in\mathcal{T}}\lambda_t\left(g_{t, \boldsymbol{\lambda}}^*(\boldsymbol{X}, S) - f_t\circ \theta(\boldsymbol{X}, S) \right)^2
\right]\enspace.
\end{equation*}
We assume in this proof that the vector $\boldsymbol{\lambda}$ is sampled from the distribution $\mathbb{P}_{\boldsymbol{\lambda}}$. Given a task $t\in\mathcal{T}$ we denote $\nu_t^* \in \argmin_{\nu}\sum_{s\in\mathcal{S}}\pi_s \mathcal{W}_2^2(\nu_{g_{t, \boldsymbol{\lambda}}^*|s}, \nu )$
where there exists $(\theta^*_t)_{t\in\mathcal{T}}$ such that $\nu_t^* = f_t \circ {\theta^*_t}$. Adapted from the work in~\cite{Chzhen_Denis_Hebiri_Oneto_Pontil20Wasser} and the universal approximation theorem \cite{hornik1989multilayer} we deduce,
\begin{multline*}\label{eq:EqMultiToUni}
    \min_{\substack{\theta \in \mathcal{H}^{\text{fair}}}} \ \mathbb{E}_{\boldsymbol{\lambda}\sim \mathbb{P}_{\boldsymbol{\lambda}}}\mathbb{E}\left[\sum_{t\in\mathcal{T}}\lambda_t\left(g_{t, \boldsymbol{\lambda}}^*(\boldsymbol{X}, S) - f_t\circ \theta(\boldsymbol{X}, S) \right)^2
\right]
    \\ = \mathbb{E}_{\boldsymbol{\lambda}\sim \mathbb{P}_{\boldsymbol{\lambda}}}\sum_{\substack{t\in\mathcal{T}\\ s\in\mathcal{S} }} \lambda_t\pi_s \mathcal{W}_2^2(\nu_{g_{t, \boldsymbol{\lambda}}^*|s}, \nu_t^*)\enspace,
\end{multline*}
which concludes the sketch of the proof, for details see the Appendix \qedsymbol{}
\end{proof}

Thm.~\ref{prop:MainFairOptimal} provides a closed form expression for the optimal fair predictor $\boldsymbol{g}_{\boldsymbol{\lambda}}^{*(\text{fair})} = \left(g^{*(\text{fair})}_{t, \boldsymbol{\lambda}}\right)_{t\in\mathcal{T}}$ for the multi-task $\boldsymbol{Y} = (Y_t)_{t\in\mathcal{T}}$. Our method is a post-processing approach, so we don't directly retrieve the parameters $\theta^{*(\text{fair})}_{\boldsymbol{\lambda}}$. 
A direct result of Thm.~\ref{prop:MainFairOptimal} indicates that our post-processing approach preserves the rank statistics~\cite{van2000asymptotic, Bobkov_Ledoux16} conditional on the sensitive feature.

\begin{corollary}[Group-wise rank statistics]\label{cor:GroupWiseRank}
     If $g_{t, \boldsymbol{\lambda}}^{*} (x_1, s) \leq g_{t, \boldsymbol{\lambda}}^{*} (x_2, s)$ for any instances $(x_1, s)$ and $(x_2, s)$ in $\mathcal{X}\times \mathcal{S}$, then the fair optimal predictor will also satisfy $g_{t, \boldsymbol{\lambda}}^{*(\text{fair})} (x_1, s) \leq g_{t, \boldsymbol{\lambda}}^{*(\text{fair})} (x_2, s)$.
\end{corollary}

To obtain the optimal fair classifier for the original two-task problem $\left(Y_1, Y_2\right)$, we can derive the final optimal fair classifier from the expression in Thm.~\ref{prop:MainFairOptimal}. Given an instance $(\boldsymbol{x}, s)\in\mathcal{X}\times \mathcal{S}$ and a threshold $\tau\in[0, 1]$, the optimal fair classifier becomes
\begin{equation*}
    c_{2, \boldsymbol{\lambda}}^{*(\text{fair})}(\boldsymbol{x}, s) = \one\left\{g_{2, \boldsymbol{\lambda}}^{*(\text{fair})}(\boldsymbol{x}, s) \geq \tau\right\}\enspace.
\end{equation*}
The finding in \cite{gaucher2022fair} is applicable to our case, where setting the threshold at $\tau=1/2$ corresponds to optimising accuracy while adhering to the DP constraint.

\section{Plug-in estimator}\label{sec:plugin}

To employ the results on real data, we propose a plug-in estimator for the optimal fair predictor $\boldsymbol{g}_{\boldsymbol{\lambda}}^{*(\text{fair})}$.

\subsection{Data-driven approach}

The estimator is constructed in two steps in a semi-supervised manner since it depends on two datasets: one labeled denoted $\mathcal{D}^{\text{train}}_n = \{(\boldsymbol{X}_i, S_i, Y_{i,1}, Y_{i,2})\}_{i=1}^{n}$ $n$ \textit{i.i.d.} copies of $(\boldsymbol{X}, S, Y_1, Y_2)$ and the other unlabeled one, denoted $\mathcal{D}^{\text{pool}}_N = \{(\boldsymbol{X}_i, S_i)\}_{i=1}^{N}$, $N$ \textit{i.i.d.} copies of $(\boldsymbol{X}, S)$. For the regression-classification problem,
\begin{itemize}
    \item[$i)$] We train \textit{simultaneously} the estimators $\widehat{g}_{1, \boldsymbol{\lambda}}$ and $\widehat{g}_{2, \boldsymbol{\lambda}}$ of respectively the regression function $g_{1, \boldsymbol{\lambda}}^{*}$ and the score function $g_{2, \boldsymbol{\lambda}}^*$ (optimal unconstrained functions) on a labeled dataset $\mathcal{D}^{\text{train}}_n$ via a multi-task learning model (see Section~\ref{sec:PbStatement}). To ensure the continuity assumption, we use a simple randomisation technique called \textit{jittering} on the predictors. For each $t\in\mathcal{T}$, we introduce
    $$
    \Bar{g}_{t, \boldsymbol{\lambda}}(\boldsymbol{X}_i, S_i, \zeta_{i,t}) = \widehat{g}_{t, \boldsymbol{\lambda}}(\boldsymbol{X}_i, S_i) + \zeta_{i,t}
    $$
    with $\zeta_{i,t}$ some uniform perturbations in $\mathcal{U}(-u, u)$ where $u$ is set by the user (e.g. $u=0.001$). This trick is often used for data visualisation for tie-breaking~\cite{chambers2018graphical, Chzhen_Denis_Hebiri_Oneto_Pontil20Wasser}. 
    The trade-off weight $\boldsymbol{\lambda}$ can be predetermined or generated during training (refer to Section~\ref{sec:empmultitask} below).
    \item[$ii)$] Empirical frequencies $\left(\widehat{\pi}_s\right)_{s\in\mathcal{S}}$, CDF $\widehat{F}_{\Bar{g}_{t, \boldsymbol{\lambda}}|s}$ and quantile function $\widehat{Q}_{\Bar{g}_{t, \boldsymbol{\lambda}}|s}$ are calibrated via the previously estimators $\Bar{g}_t$ and the unlabeled data set $\mathcal{D}^{\text{pool}}_N$.
\end{itemize}
The \textit{(randomised) Wasserstein fair estimator} for each $t\in \mathcal{T}$ is defined by plug-in
\begin{equation}\label{eq:FairPredEmp}
    \widehat{g}^{(\text{fair})}_{t, \boldsymbol{\lambda}}(\boldsymbol{x}, s) = \sum_{s'\in\mathcal{S} } \widehat{\pi}_{s'}\widehat{Q}_{\Bar{g}_{t, \boldsymbol{\lambda}}|s'} \circ \widehat{F}_{\Bar{g}_{t, \boldsymbol{\lambda}}|s}\left(\Bar{g}_{t, \boldsymbol{\lambda}}(\boldsymbol{x}, s,\zeta_t)\right)
\end{equation}
with $(\zeta_t)_{t\in \mathcal{T}}\overset{i.i.d.}{\sim} \mathcal{U}(-u, u)$. We present the associated pseudo-code in Alg.\ref{alg:optimization}. 

\begin{remark}[Data splitting]
    The procedure requires unlabeled data. If we do not have any in practice, we can split the labeled data in two and remove the labels in one of the two sets. As demonstrated in~\cite{denis2021fairness}, splitting the data is essential to avoid overfitting and to get the right level of fairness.
\end{remark}

\begin{algorithm}
   \caption{Fairness calibration}
   \label{alg:optimization}
\begin{algorithmic}
   \STATE {\bfseries Input:} new data point $({x}, {s})$, base estimators $(\Hat{g}_{t, \boldsymbol{\lambda}})_{t\in\mathcal{T}}$, unlabeled sample $\mathcal{D}^{\text{pool}}_N$, and  \emph{i.i.d} uniform perturbations $(\zeta_{k,i}^{s})_{k,i,s}$.
   \STATE {\bf \quad Step 0.} Split $\mathcal{D}^{\text{pool}}_N$ to construct $(S_i)_{i=1}^N$ and $\{X_{i}^s\}_{i=1}^{N_s}\sim \mathbb{P}_{X|S=s}$ given $s\in \mathcal{S}$;
   \STATE {\bf \quad Step 1.} Compute the empirical frequencies $(\hat{\pi}_s)_s$ based on $(S_i)_{i=1}^N$;
   \STATE {\bf \quad Step 2.} Compute the empirical CDF $\widehat{F}_{\Bar{g}_{t, \boldsymbol{\lambda}}|s}$ and quantile $\widehat{Q}_{\Bar{g}_{t, \boldsymbol{\lambda}}|s'}$ from $\{X_{i}^s\}_{i=1}^{N_s}$;
   \STATE {\bf \quad Step 3.} Compute $\hat{g}_{1, \boldsymbol{\lambda}}, \ldots, \hat{g}_{|\mathcal{T}|, \boldsymbol{\lambda}}$ thanks to Eq.\eqref{eq:FairPredEmp};
   \STATE {\bfseries Output:} fair predictors $\hat{g}_{1, \boldsymbol{\lambda}}(\boldsymbol{x},s), \ldots, \hat{g}_{|\mathcal{T}|, \boldsymbol{\lambda}}(\boldsymbol{x},s)$ at point $(\boldsymbol{x},s)$.
\end{algorithmic}
\end{algorithm}

\subsection{Empirical Multi-Task}\label{sec:empmultitask}
This section outlines how we build each marginal predictor $\Hat{g}_{t, \boldsymbol{\lambda}}$ using the training set $\mathcal{D}^{\text{train}}_n = (\boldsymbol{x}_i, s_i, \boldsymbol{y}_i)_{i=1}^{n}$ where each $(\boldsymbol{x}_i, s_i, \boldsymbol{y}_i)$ is a realisation of
$(\boldsymbol{X}_i, S_i, \boldsymbol{Y}_i) \sim \mathbb{P}$. Given a set of task-related loss functions $\mathcal{L}_t$, we define the empirical multi-task problem from Eq.\eqref{eq:multitask} as 
 \begin{equation*}\label{eq:empirical_multitask}
     \hat{\boldsymbol{\theta}}_\lambda = \argmin_{\theta}\sum_{i=1}^n\sum_{t=1}^2 \lambda_t \mathcal{L}_t(y_{i,t}, f_t \circ {\theta}(\boldsymbol{x}_i, s_i)).
\end{equation*}
As the values for different loss functions $\mathcal{L}_t$ are situated on different scales, issues arise during training when using gradient based methods (see for example \cite{yu2020gradient, wanggradient,liu2021conflict,navon_bargaining_22a} for discussions about the issue). The $\boldsymbol{\lambda}$ parameter can alleviate this issue but is difficult to find in practice. Since there is no a priori optimal choice, we use the \emph{"You Only Train Once"} (YOTO) approach of \cite{dosovitskiy2020you}, initially developed for regression-regression problems. As the name of their approach suggests, the model is only trained once for a host of different $\boldsymbol{\lambda}$ values by conditioning the parameters of the neural network directly on the task weights $\boldsymbol{\lambda}$. The key idea is that different values for $\boldsymbol{\lambda}$ are sampled from a distribution and included directly in the estimation process. Rewritten, Eq.\eqref{eq:empirical_multitask} then becomes:
\begin{equation}
    \hat{\boldsymbol{\theta}}_{\boldsymbol{\lambda}} = \argmin_{\theta}\sum_{i=1}^n\sum_{t=1}^2 \lambda_t \mathcal{L}_t(y_{i,t}, f_t \circ \theta(\boldsymbol{x}_i, s_i; \boldsymbol{\lambda})), \quad \boldsymbol{\lambda} \sim \mathbb{P}_{\boldsymbol{\lambda}}
\end{equation}
where $\mathbb{P}_{\boldsymbol{\lambda}}$ is a sampling distribution. For our purposes, we use uniform distribution. As in the original article \cite{dosovitskiy2020you}, we employ FiLM conditioning developed by \cite{perez2018film} to condition each layer of $\theta(\cdot)$ directly on the sampled $\boldsymbol{\lambda}$. Once the model is fitted, the optimal $\boldsymbol{\lambda}$ is chosen via a problem specific calibration method on a calibration set. Precise details on the implementation can be found in Alg.~\ref{alg:optimization2}.
\begin{algorithm}
   \caption{$\boldsymbol{\lambda}$-calibrated MTL}
   \label{alg:optimization2}
\begin{algorithmic}
   \STATE {\bfseries Input:} Training data $\mathcal{D}_n^{\text{train}}$, bounds $b_l, b_u$ for $\mathcal{U}(b_l, b_u)$, model, validation grid
   \WHILE{training}
        \STATE {\bf \quad Step 1.} Draw $n_b$ $\lambda_t \sim \mathcal{U}(b_l, b_u)$;
        \STATE {\bf \quad Step 2.} FiLM Condition\cite{perez2018film} each layer in neural network using $\boldsymbol{\lambda}$;
        \STATE {\bf \quad Step 3.} Condition loss as in YOTO \cite{dosovitskiy2020you} $t$ with $\lambda_t$;
        \STATE {\bf \quad Step 4.} Adjust model parameters given $x,s,\boldsymbol{\lambda}$;
   \ENDWHILE
   \FOR{$\boldsymbol{\lambda}_v$ in validation grid}
        \STATE {\bf \quad Step 1.} Predict $\hat{y_t}$ for all $t$ with $x,s,\boldsymbol{\lambda}_v$;
        \STATE {\bf \quad Step 2.} Evaluate $\hat{y_t}$, $y_t$ for all $t$
   \ENDFOR
   \STATE {\bfseries Output:} Grid of task-wise error metrics given all $\boldsymbol{\lambda}_v$ in validation grid, choose optimal $\boldsymbol{\lambda}_v$
\end{algorithmic}
\end{algorithm}

\section{Numerical evaluation}\label{sec:application}

To evaluate the numerical performance, we conduct experiments on different datasets\footnote{All sourcecode and data links can be found on \href{https://github.com/phi-ra/FairMultitask}{\texttt{github.com/phi-ra/FairMultitask}}}. All data sets used are publicly available and are described in the next subsection. We also describe each of the separate tasks and the variable on which we want to achieve demographic parity (the $S$ in the equations above). 

\subsection{Datasets}

We focus on applications with tabular data, the first data set we consider stems from the \textsc{folktables} package \cite{ding2021retiring}, which was constructed to enable bench marking of machine learning models\footnote{\href{https://github.com/socialfoundations/folktables}{\texttt{github.com/socialfoundations/folktables}}}. Instead of a single task, we consider the simultaneous prediction of both \emph{Mobility} (Binary) and \emph{Income} (Regression) using a set of 19 features. Here, we consider \emph{gender} the binary sensitive variable. In total, we use 58,650 observations from the state of California.

As a second benchmark, we consider the \textsc{compas} data set \cite{larson_angwin_kirchner_mattu_2016}. It was constructed using a commercial algorithm which is used to assess the likelihood of reoffending for criminal defendants. It has been shown that its results are biased in favour of white defendants, and the data set has been used to assess the efficacy of other fairness related algorithms \cite{OnetoC062}\footnote{Although available publicly, we believe the usage of the data needs to undergo some ethical considerations. Please read our separate ethical statement regarding this}. The data set collected has two classification targets (\emph{recidivism} and \emph{violent recidivism}), that are predicted using 18 features. In total, we use 6,172 observations from the data set and, in the spirit of the initial investigation, we consider \emph{race} as the sensitive attribute. 

\subsection{Methods}

For the simulations, we split data into 80/20 train/test set. All estimators are based on neural networks with a fixed architecture and 10\% dropout in the layers. We compare the performance and fairness of the optimal predictor and the optimal fair predictor across a MTL model and two single-task (STL) models, across 20 bootstrap iterations. We refrain from an in-depth architecture and hyper-parameter search to keep the insights comparable among the simulations. 

Our goal is to exemplify two distinct features of MTL under fairness constraints. A standard application in MTL is to leverage similarities in tasks to improve performance in the case where labels in one of the tasks are scarce. As our method is valid for any trade-off weight $\boldsymbol{\lambda}$, we can achieve fairness even in the case where one task is more important than the other. To simulate this environment, we successively remove [0,25,50,75,95]\% of the regression labels in the training of the \textsc{folktables} data set and calibrate the $\boldsymbol{\lambda}$ vector to optimise performance on the regression task. Intuitively, we would expect the predictive performance of the models to degrade with a higher proportion of missing data, but MTL should perform better than STL, if it is able to extract knowledge from the related classification task. A second use for MTL arises when we are interested in the joint distribution of several tasks. This is of particular importance for the second case, as one of the tasks in the \textsc{compas} data set is actually a subset of the other. To illustrate this, we optimise the $\boldsymbol{\lambda}$ parameter for the \textsc{compas} tasks in order to maximise performance in both. To measure the performance we use the mean-squared error (MSE) of the log-predictions for the regression task and area under the ROC curve (AUC) for the classification tasks. To calculate the unfairness, we compare the predictions made on the two sub-populations specified by the presence (\emph{Protected}) or absence (\emph{Unprotected}) of the sensitive attribute using the empirical counterpart $\Hat{\mathcal{U}}(g_t)$ of the unfairness given in Definition~\ref{eq:Unfairness} which corresponds to a two-sample Kolmogorov-Smirnov (KS) test
\begin{equation*}
\Hat{\mathcal{U}}(g_t) := \sup\limits_{u\in\mathcal{Y}_t} \left|\ \hat{F}_{g_t|1}(u) - \hat{F}_{g_t|-1}(u)\ \right| \enspace.
\end{equation*}

\begin{figure}[t!]
\centering
\begin{minipage}{.5\textwidth}
  \centering
  \vspace{0.2cm}
  \includegraphics[width=.82\linewidth]{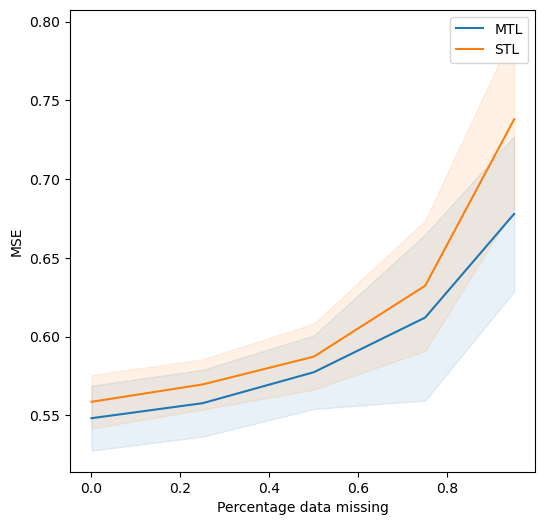}
  \label{fig:test1}
\end{minipage}%
\begin{minipage}{.5\textwidth}
  \centering
  \includegraphics[width=.82\linewidth]{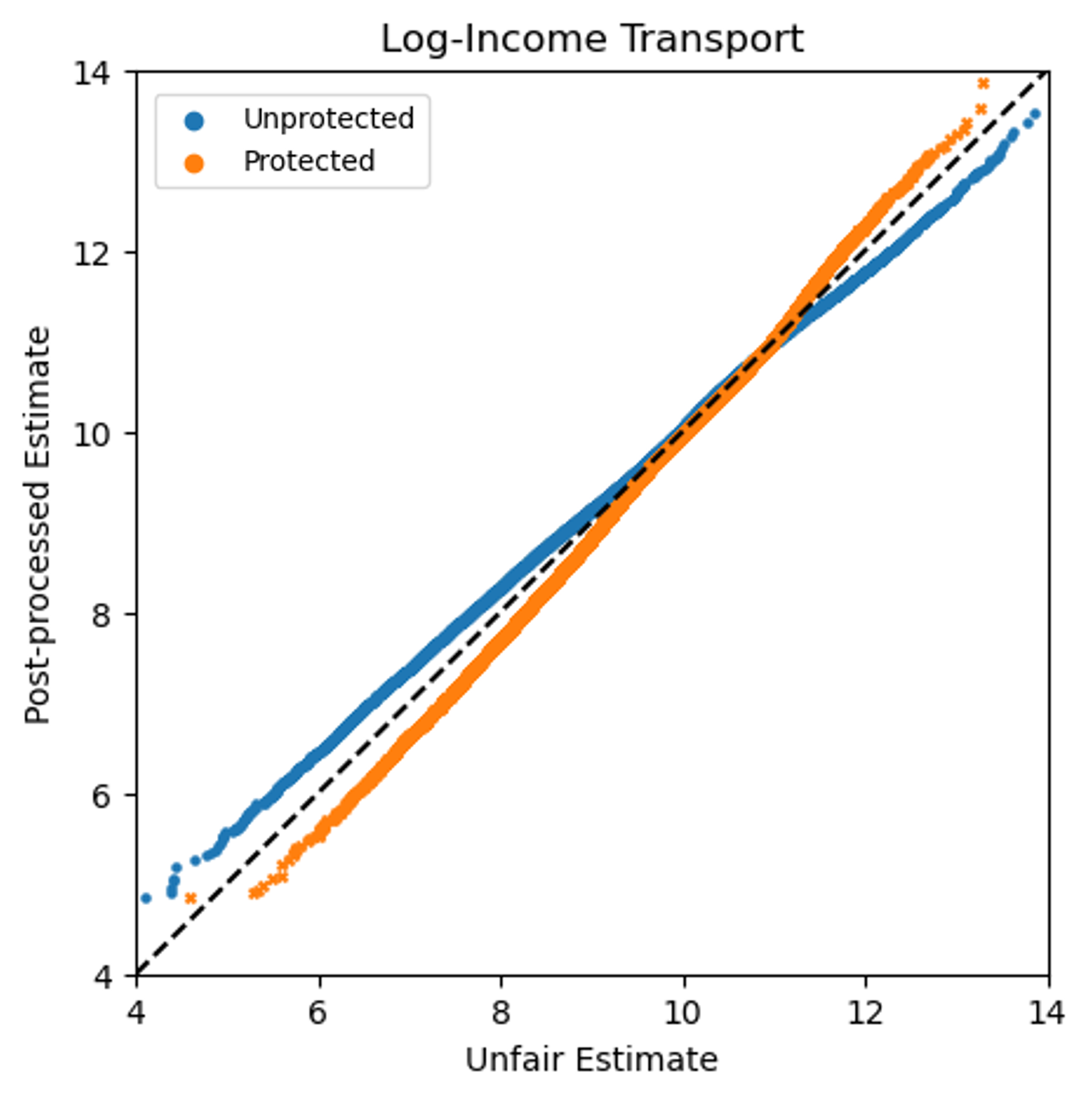}
\end{minipage}
\caption{Left, the performance as measured by MSE for MTL and STL, here the $\boldsymbol{\lambda}$ parameter was chosen to optimise the regression task. This leads to better outcomes, especially in the case of missing values in the regression labels. Right, regression estimates before versus after the optimal transport. }
\label{fig:highlights_folk}
\end{figure}

\subsection{Results}

The numeric results for the \textsc{folktables} data set are summarised in Table \ref{table:resultsfolk} and highlights visualised in Figure \ref{fig:highlights_folk}. Especially the \emph{Income} variable (the regression task) suffers from unfairness (as indicated by a higher value in the KS test). The advantage of using a second task to help the predictions is also clearly visible in the numerical results and the left pane of Figure \ref{fig:highlights_folk}. Although the performance of MTL deteriorates with more missing labels, it suffers less than the STL estimation. The classification task performs less well, as the $\boldsymbol{\lambda}$ was calibrated to optimise the regression task. Additionally, as there are no missing labels in the classification task, we would expect only marginal gains from using MTL even in the case where $\boldsymbol{\lambda}$ is calibrated to serve both tasks well. This is in line with what was found in the literature of MTL \cite{standley2020tasks}. Here, the specification using the YOTO approach allows the user to chose the optimal trade-off weight for the problem at hand in a specific calibration step which will lead to different outcomes using the same trained weights. The advantage of our result is that it will be valid for any $\boldsymbol{\lambda}$. We can also see across the board that the imposing fairness among the predictions reduces slightly the predictive performance and almost exactly satisfies the DP condition. We also visualise the effect of the optimal transport as specified by the Wasserstein fair estimator in Eq.\eqref{eq:FairPredEmp}, as suggested in \cite{charpentier_book}. Because our operations preserve the group-wise rank (Cor.~\ref{cor:GroupWiseRank}), we can directly represent the changes in the predictions for each group. The predicted income distribution is shifted in a way such that the upper tail for the sensitive group is shifted up, but the lower tail is shifted downwards. 
\begin{table*}[t!]
\begin{normalsize}
    \resizebox{\linewidth}{!}{
    \begin{tabular}{|l||*{6}{c|}} \hline
    \multirow{2}{*}{\begin{tabular}{p{0.5cm}}\backslashbox[30mm]{Data}{Model} 
    \end{tabular}} & \multicolumn{2}{c|}{MTL} & \multicolumn{2}{c|}{MTL, Post-processed} & \multicolumn{2}{c|}{STL} \\ \cline{2-7}
     & Performance & Unfairness & Performance & Unfairness & Performance & Unfairness \\ \hline\hline
    regression - all data & 
    $0.548 \pm 0.02$ & 
    $0.109 \pm 0.01$ &
    $0.558 \pm 0.02$ &
    $0.018 \pm 0.00$ &
    $0.559 \pm 0.02$ &
    $0.107 \pm 0.01$  \\
    regression - 25\% missing &
    $0.558 \pm 0.02$ & 
    $0.109 \pm 0.02$ & 
    $0.572 \pm 0.02$ & 
    $0.018 \pm 0.00$ & 
    $0.570 \pm 0.02$ & 
    $0.105 \pm 0.02$ \\
    regression - 50\% missing &
    $0.577 \pm 0.02$ & 
    $0.109 \pm 0.02$ & 
    $0.593 \pm 0.03$ & 
    $0.018 \pm 0.01$ & 
    $0.587 \pm 0.02$ & 
    $0.099 \pm 0.01$ \\
    regression - 75\% missing &
    $0.612 \pm 0.05$ & 
    $0.101 \pm 0.02$ & 
    $0.627 \pm 0.06$ & 
    $0.019 \pm 0.01$ & 
    $0.632 \pm 0.04$ & 
    $0.098 \pm 0.01$ \\
    regression - 95\% missing &
    \cellcolor{blue!15}$0.678 \pm 0.05$ & 
    $0.105 \pm 0.02$ & 
    \cellcolor{blue!15}$0.687 \pm 0.05$ & 
    $0.018  \pm 0.01$ & 
    $0.738 \pm 0.06$ & 
    $0.108 \pm 0.03$ \\
    \hline
    classification - all data & 
    $0.576 \pm 0.01$ & 
    $0.080  \pm 0.03$ & 
    $0.577 \pm 0.01$ & 
    $0.018 \pm 0.01$ & 
    \cellcolor{blue!15}$0.640 \pm 0.03$ & 
    $0.042 \pm 0.02$  \\
    \hline
    \hline
    \end{tabular}}
    \vspace{0.1cm}
    \caption{Performance and unfairness for MTL and Single Task Learning (STL) models on the \textsc{folktables} data. Each model was also post-processed and evaluated on performance and unfairness.}
    \label{table:resultsfolk}
    \end{normalsize}
\end{table*}

\begin{figure}
\centering
\includegraphics[width=0.8\textwidth]{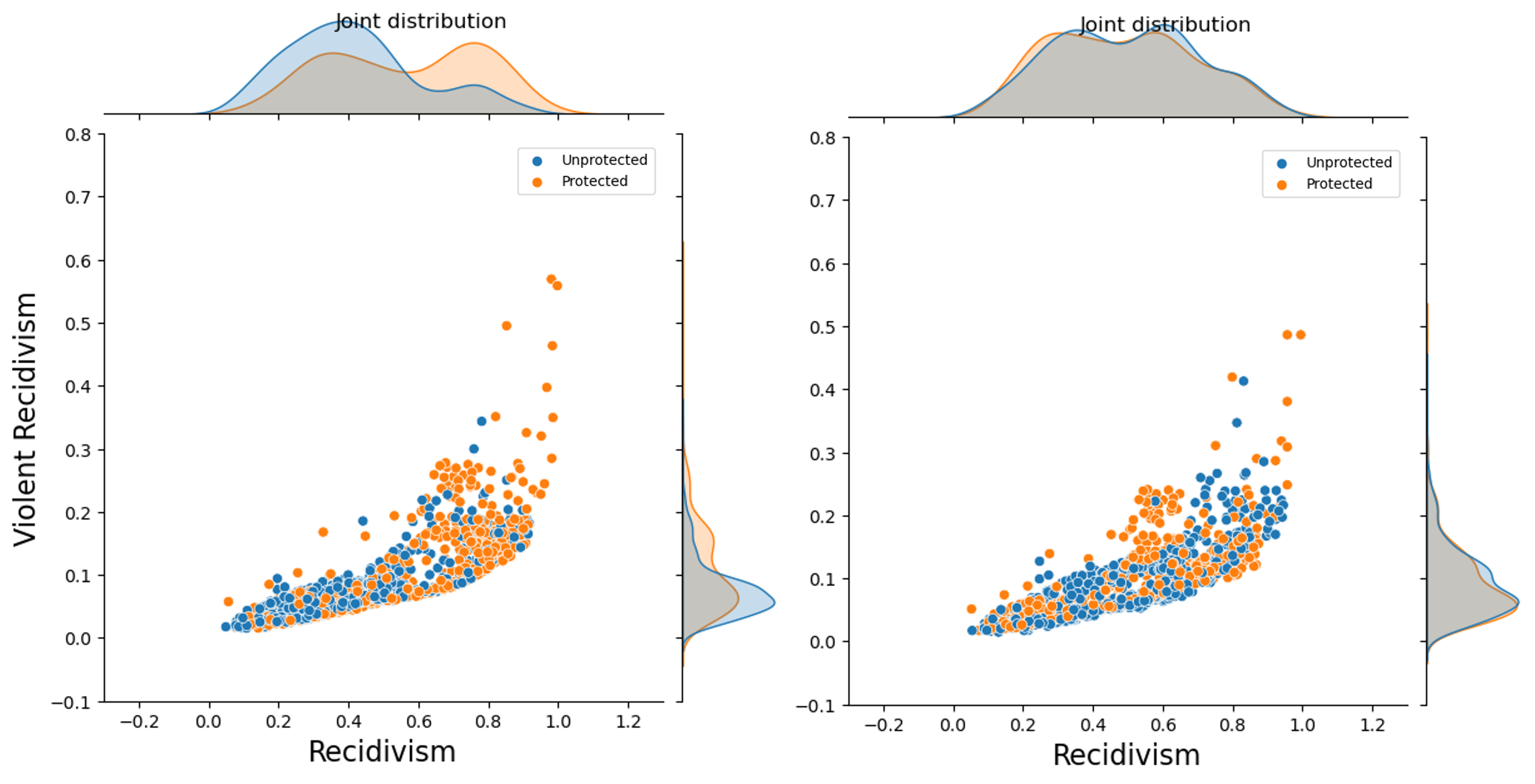}
\caption{Joint distribution for scores under unconstrained and DP-fair regimes. Color indicates the presence of the sensitive feature. Note that the joint distribution appears more mixed and the marginal distributions overlap in the DP fair case. } \label{fig:fairness_compas}
\end{figure}

The results from the \textsc{compas} data set mirror in large parts the ones of the \textsc{folktables} but here we want to optimise the performance across both tasks at once. Results are summarised in Table \ref{table:resultscompas} and visualised in Figure \ref{fig:fairness_compas}. 
The effect of the optimal transport on the distributions can be seen in the marginal distributions in \ref{fig:fairness_compas}. The colors indicate whether a given individual is identified as belonging to a protected group. Clearly a bias can be seen in the marginal distributions, the protected group has both a higher recidivism score and a slightly higher violent recidivism score, which mirrors the findings from \cite{larson_angwin_kirchner_mattu_2016}. In the right pane, we show the post-processed version, where the marginal distributions are almost congruent, enforcing the DP condition. The resulting fairness is also assessed numerically using the KS test. As expected this also leads to a small performance decrease as measured by AUC. The tuning of the $\boldsymbol{\lambda}$ parameter allows to have a predictive performance that is almost equivalent to the STL specification, with the advantage that we can jointly predict the scores and enforce the DP condition for this joint representation.
\begin{table*}[h!]
\begin{normalsize}
    \resizebox{\linewidth}{!}{
    \begin{tabular}{|l||*{8}{c|}} \hline
    \multirow{2}{*}{\begin{tabular}{p{0.5cm}}\backslashbox[30mm]{Data}{Model} 
    \end{tabular}} & \multicolumn{2}{c|}{MTL} & \multicolumn{2}{c|}{MTL, Post-processed} & \multicolumn{2}{c|}{STL} &
    \multicolumn{2}{c|}{STL, Post-processed}\\ \cline{2-9}
     & Performance & Unfairness & Performance & Unfairness & Performance & Unfairness & Performance & Unfairness \\ \hline\hline
    task 1 - all data & 
    $0.742 \pm 0.01$ & 
    $0.289 \pm 0.02$ &
    $0.727 \pm 0.01$ &
    $0.052 \pm 0.02$ &
    $0.745 \pm 0.01$ &
    $0.291 \pm 0.02$ &
    $0.730 \pm 0.01$ & 
    $0.055 \pm 0.02$ \\
    \hline
    task 2 - all data & 
    $0.686 \pm 0.02$ & 
    $0.289 \pm 0.04$ & 
    $0.649 \pm 0.01$ & 
    $0.053 \pm 0.02$ & 
    $0.671 \pm 0.01$ & 
    $0.290 \pm 0.03$ & 
    $0.638 \pm 0.03$ & 
    $0.053 \pm 0.02$ \\
    \hline
    \hline
    \end{tabular}}
    \vspace{0.1cm}
    \caption{Performance in AUC and unfairness for MTL and Single Task Learning (STL) models on the \textsc{compas} data. Each model was also post-processed and evaluated on performance and unfairness.}
    \label{table:resultscompas}
    \end{normalsize}
\end{table*}

\section{Conclusion}\label{sec:conclusion}

As multi-task learning grows in popularity, ensuring fairness among the predictions becomes a new challenge as the precise effects of MTL are still poorly understood. In this paper, we investigated the general effects of parameter sharing on the marginal tasks. We proposed a method to integrate fairness into MTL through a post-processing procedure which keeps a key advantage of MTL, shorter computational expenses, largely intact. This also opens a host of new directions for further research. As we focused on tabular data, we were less restricted by possible model architectures. In other related areas where MTL is becoming more popular, such as computer vision, pre-trained models akin to our $h_{\theta}$ are often used to ease the computational burden. A thorough investigation into the precise effects of the combination of the triple Transfer-Multitask-Fair learning would hence be a natural extension. A further extension of our results would be to consider fairness in a general multivariate setting. This would mean shifting the parameters of the embedding $h_{\theta}$ simultaneously for all tasks. This will likely not be possible with a similar closed-form solution, as our approach relies on the estimation of quantiles. As MTL is generally used in the case where there is a rather strong (and exploitable) relationship between the two tasks, the marginal approach we propose here seems apt, but a theoretical discussion would nevertheless be interesting.



\section*{Ethics statement}

Our work is centered around fairness, which is a goal we sincerely believe all model should strive to achieve. Nevertheless, to ensure fairness in models, one needs to define unfairness as its counterpart. This naturally leads to a conundrum when performing research on this topic. On one hand, we would like our models to be fair, but to analyse the differences and show an improvement, we first need to create an unfair outcome. As has been shown in the past, simply ignoring the sensitive attributes does not solve the problem of bias in the data. Further, as more flexible methods make their way into practical modelling, this issue is only bound to increase. Hence it is our conviction that estimating intentionally unfair models (by for example including sensitive variables explicitly in the training phase) is ethically justifiable if the goal is to provide a truly fair estimation. In that sense our work contributes to achieving fairness, and does not create new risks by itself. 

    In our empirical application, we consider data which was used in a predictive algorithm in the criminal justice system. This is particularly concerning as there have been numerous instances where racial, ethnic or gender bias was detected in such systems (indeed the data from \textsc{compas} were collected to show precisely that) and the criminal justice system is supposed to be egalitarian. Further, existing biases within the justice system may be further reinforced. Although the above mentioned weaknesses are well documented, such algorithms continue to be used in practice. Our work does not contribute to these algorithms directly but rather uses them as an example to show unequal treatment. Whereas the usage of other, biased data sets, such as the well-known \emph{Boston Housing} data set is discouraged, we believe that in order to show the effectiveness of fairness related algorithms, the use of such a data set is justified.

%
%

\clearpage

\bibliographystyle{splncs04}
\bibliography{biblio}

\end{document}